\documentclass{article}

\PassOptionsToPackage{numbers, compress}{natbib}

\usepackage[final]{neurips_wrl2023}

\usepackage[utf8]{inputenc} 
\usepackage[T1]{fontenc}    
\usepackage[hidelinks]{hyperref}       
\usepackage{url}            
\usepackage{booktabs}       
\usepackage{amsfonts}       
\usepackage{nicefrac}       
\usepackage{microtype}      
\usepackage{graphicx}
\usepackage{glossaries}
\usepackage{todonotes}

\title{LocoMuJoCo: A Comprehensive Imitation Learning \\
Benchmark for Locomotion}

\author{%
  Firas Al-Hafez$^1$ $\quad$ Guoping Zhao$^{2}$ $\quad$ Jan Peters$^{1, 3, 4, 5}$ $\quad$ Davide Tateo$^1$\\
    $^1$Intelligent Autonomous Systems Group, TU Darmstadt \\
    $^2$Locomotion Laboratory, TU Darmstadt\\
     $^3$German Research Center for AI (DFKI)\\
     $^4$Centre for Cognitive Science\\
     $^5$Hessian.AI\\
  \texttt{\{name.surname\}}\texttt{@tu-darmstadt.de} \\
}

\newacronym{rl}{RL}{Reinforcement Learning}
\newacronym{drl}{DRL}{Deep Reinforcement Learning}
\newacronym{il}{IL}{Imitation Learning}
\newacronym{irl}{IRL}{Inverse Reinforcement Learning}
\newacronym{maxent}{MaxEnt}{Maximum Entropy}

\newacronym{avi}{AVI}{Approximate Value-Iteration}
\newacronym{api}{API}{Approximate Policy-Iteration}
\newacronym[plural=MDPs, firstplural=Markov Decision Processes (MDPs)]{mdp}{MDP}{Markov Decision Process}
\newacronym{cmdp}{CMDP}{Constrained Markov Decision Processes}
\newacronym{kl}{KL}{Kullback-Leibler Divergence}
\newacronym{gae}{GAE}{Generalized Advantage Estimation}
\newacronym{trpo}{TRPO}{Trust Region Policy Optimization}
\newacronym{gans}{GANs}{Generative Adversarial Networks}
\newacronym{ail}{AIL}{Adversarial Imitation Learning}
\newacronym{gail}{GAIL}{Generative Adversarial Imitation Learning}
\newacronym{vail}{VAIL}{Variational Adversarial Imitation Learning}
\newacronym{gaifo}{GAIfO}{Generative Adversarial Imitation
from Observation}
\newacronym{lsgail}{LS-GAIL}{Least-Squares Generative Adversarial Imitation Learning algorithms}
\newacronym{sqil}{SQIL}{Soft Q-Imitation Learning}
\newacronym{iq}{IQ-Learn}{Inverse soft Q-Learning}
\newacronym{sac}{SAC}{Soft-Actor Critic}
\newacronym{lsiq}{LS-IQ}{Least Squares Inverse Q-Learning}
\newacronym{lsgans}{LSGANs}{Least Squares Generative Adversarial Networks}
\newacronym{idm}{IDM}{Inverse-Dynamics Model}
\newacronym{pomdp}{POMDP}{Partially Observable Markov Decision Process}
\newacronym{mpc}{MPC}{Model Predictive Control}

\glsdisablehyper

\begin{document}

\maketitle

\begin{abstract}
\gls{il} holds great promise for enabling agile locomotion in embodied agents. However, many existing locomotion benchmarks primarily focus on simplified toy tasks, often failing to capture the complexity of real-world scenarios and steering research toward unrealistic domains. To advance research in \gls{il} for locomotion, we present a novel benchmark designed to facilitate rigorous evaluation and comparison of \gls{il} algorithms. This benchmark encompasses a diverse set of environments, including quadrupeds, bipeds, and musculoskeletal human models, each accompanied by comprehensive datasets, such as real noisy motion capture data, ground truth expert data, and ground truth sub-optimal data, enabling evaluation across a spectrum of difficulty levels. To increase the robustness of learned agents, we provide an easy interface for dynamics randomization and offer a wide range of partially observable tasks to train agents across different embodiments. Finally, we provide handcrafted metrics for each task and ship our benchmark with state-of-the-art baseline algorithms to ease evaluation and enable fast benchmarking. The code and videos can be found here:\\
\url{https://github.com/robfiras/loco-mujoco}.
\end{abstract}

\glsresetall

\section{Introduction}
\gls{il} plays a pivotal role in machine learning, especially in robotics, enabling rapid skill acquisition without requiring manual skill programming or reward function tuning. Recent advancements, particularly in Adversarial \gls{il} and \gls{irl}, have expanded the applicability of \gls{il} algorithms to complex, high-dimensional tasks. In locomotion, \gls{il} is essential due to the inherent complexity of defining reward functions and limitations of classical approaches in challenging environments. However, the lack of standardized benchmarks and datasets is a significant issue in the field. Many \gls{il} works use locomotion tasks for evaluation but define their own unique environments and expert datasets. This variability hampers evaluation consistency, conceals method strengths and weaknesses, and impedes reproducibility. While many \gls{il} benchmarks are available, particularly for manipulation, we still lack a proper benchmark for the locomotion tasks. Existing benchmarks for this setting either tackle toy tasks, are very far from realistic settings and real robotic platforms or are not designed for \gls{il}, therefore lacking proper datasets. 


To address this challenge, we present LocoMuJoCo, a Python-based benchmark tailored for locomotion within the context of imitation learning (\gls{il}). Our benchmark is designed to enhance method usability by offering compatibility with Gymnasium~\cite{towers_gymnasium_2023}, a widely adopted interface for \gls{rl} algorithms. Additionally, we provide native support for the Mushroom-RL~\cite{mushroomrl} library. Furthermore, we supply a collection of baseline approaches capable of yielding reasonable solutions for the benchmark tasks. LocoMuJoCo encompasses various locomotion tasks, including humanoid locomotion, quadruped locomotion, and musculoskeletal human models. For each of these environments, we provide a wide variety of tasks and comprehensive datasets, including a) realistic noisy motion capture data mapped to the respective embodiment, b) ground truth expert data with actions for many environments, and c) ground truth sub-optimal expert data with actions for regularization in offline \gls{irl} or preference-based \gls{il}. This wide variety of tasks and datasets allows the benchmark to cover a broad spectrum of complexities, ranging from relatively simple tasks to extremely challenging problems yet to be fully solved. To propel application to real-world robotic systems, LocoMuJoCo allows the users to easily include domain randomization, reducing the sim-to-real gap. Finally, our benchmark provides a reward function for every task, allowing the user to measure the agent's performance. Alternatively, the user can define his own reward function to use it either as a performance metric or to use the whole benchmark for pure \gls{rl}.
%
%
\paragraph{Related Work.}
RLBench \cite{james2022} is a benchmarking platform for testing and evaluating \gls{rl} algorithms in the context of robotic manipulation. It offers standardized tasks and environments and provides a planner to solve each task, allowing the use of this benchmark for \gls{il}. It comes with a realistic Franka Panda robot arm model, allowing zero-shot transfer of the policies to the real world \cite{alhafez2021}. Meta-World \cite{yu2019meta} a benchmark for multi-task and meta \gls{rl}. However, it does not provide expert demonstration, rendering it unsuitable for imitation learning. Robosuite \cite{robosuite2020} is a  manipulation benchmark focusing on modularity, which provides utilities for collecting human demonstrations for \gls{il}. Simitate \cite{memmesheimer2019} is a benchmark explicitly designed for \gls{il}, where a dataset of 1938 sequences of human tasks are provided together with a simulator. Franka Kitchen \cite{abhishek2019} is another multi-task manipulation benchmark based on the Franka Panda robot, in which multiple tasks have to be solved in parallel. While an unlabelled dataset is available, it is not suitable for \gls{il}. Adroit \cite{rajeswaran2018} is a small manipulation benchmark using a floating human hand, which also comes with a set of demonstrations. Gym-Mujoco \cite{openaigym2016} and DeepMind Control Suite \cite{tunyasuvunakool2020} are two popular reinforcement benchmarks encompassing many locomotion tasks, ranging from 2D-walker to simple humanoid tasks. Both of the benchmarks do not explicitly focus on the \gls{il} setting and only consider toy tasks with dynamics fine-tuned for \gls{rl}. D4RL \cite{fu2020d4rl} is an offline \gls{rl} benchmark, including datasets for, inter alia, Franka Kitchen, Adroit, Gym-Mujoco, and DeepMind Control Suite. For each task, multiple datasets represent agents of different performances, which can be used for \gls{il}. Finally, the Myosuite~\cite{caggiano22a} is a \gls{rl} benchmark for general muscle-actuated systems, including a humanoid. However, the benchmark does not provide expert data for imitation learning. None of the aforementioned benchmarks focuses on \gls{il} for realistic locomotion, emphasizing the need for such a benchmark. 
\section{Benchmark Design}
%
In this section, we introduce the framework of our benchmark, which has been meticulously crafted for versatility. This benchmark accommodates various scenarios of \gls{il}, such as cases involving \gls{il} without access to expert actions or dealing with correspondence mismatches of different embodiments. It encompasses tasks spanning a range of difficulty levels and offers user-friendliness, effectively addressing the diverse requirements of \gls{il} in realistic locomotion scenarios.
%
%
\begin{figure}
    \centering
    \includegraphics[width=.95\textwidth]{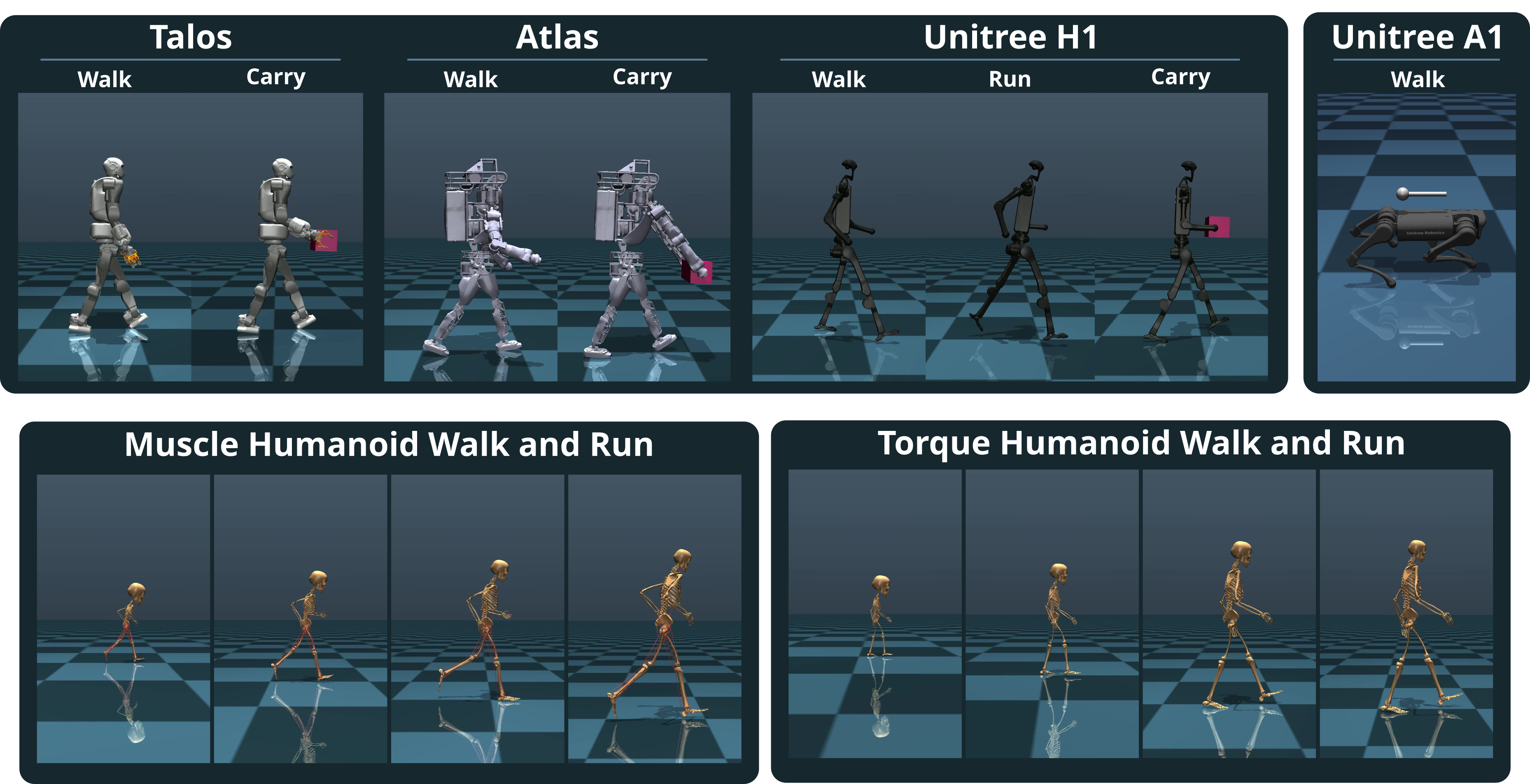}
    \caption{Overview of environments. Each task is defined by a certain dataset in an environment, e.g., the Talos carry or the muscle humanoid running task. Currently, LocoMuJoCo encompasses 12 environments with a total of 27 tasks. }
    \label{fig:overview_envs}\vspace{-0.3cm}
\end{figure}

\textbf{Environments} are the core building blocks of this benchmark and are presented in Figure \ref{fig:overview_envs}. At the time of writing, the benchmark includes 12 environments. The environments consist of 6 base environments: the Talos robot, the Atlas robot, the Unitree H1 humanoid, the Unitree A1 quadruped, a torque-actuated skeleton model, and a musculoskeletal human model. Additionally, both human models can be scaled to 4 different human sizes to represent an infant ($\sim$ 2-year-old), a child ($\sim$ 5-year-old), a teenager ($\sim$ 12-year-old), and an adult. To generate the different models, the links (i.e., the bones) are scaled linearly, the masses are scaled cubically, the inertias are scaled quintically, and the actuator torques or muscle forces are scaled quadratically w.r.t. the height of the human model. The skeleton models are converted models from \cite{HAMNER20102709}, the Talos model is converted from the official library\footnote{https://github.com/pal-robotics/talos\_robot}, the Atlas model is converted from the one provided by the Open Source Robotics Foundation\footnote{https://github.com/osrf/drcsim}, and the Unitree H1 and A1 models are taken from the MuJoCo menagerie \cite{menagerie2022github}.

\textbf{Tasks} are defined by a dataset representing the expert's behavior. For example, the musculoskeletal human environment encompasses two tasks, walking and running, represented by two datasets. Each environment comes with at least two tasks, resulting in 27 tasks in total, with more to come in the future. Each dataset can be quickly loaded and replayed to allow easy inspection. Furthermore, we construct \gls{pomdp} environments by randomizing the carry weight in the Atlas, Talos, and the Unitree H1 environment or randomizing the height of the skeleton model. In these cases, state masks are provided, which can be used to hide the important information regarding these tasks -- weight of mass or type of humanoid. In doing so, the user can choose which part of his algorithm can get access to privileged information and which not. A typical use case is the teacher-student scenario, in which the policy only receives an observation -- the masked state -- and the critic gets the complete state information. Finally, we provide the possibility of initializing the simulation from either a randomly sampled or a specifically chosen expert demonstration, which has proven crucial in imitation learning settings with environment interactions \cite{peng2018, peng2021}. 

\textbf{Datasets} not only express the tasks that are supposed to be solved but are also a tool to accommodate various scenarios of \gls{il} defining a broad spectrum of difficulties. Within this benchmark, we cover the following \gls{il} paradigms: a) learning with/without access to expert actions; b) learning under mismatches between the expert's and agent's embodiments; c) learning with sub-optimal expert states and actions. The first paradigm is of particular importance since it is often easy to access an expert's observation -- e.g., the kinematic trajectory of a human walking -- but it is often impossible to access its actions -- the underlying muscle actuation. The second paradigm is essential since it is easy to collect data for some embodiment -- human running -- but often difficult or even impossible under some other -- Atlas running. Hence, we provide realistic motion capture datasets for all environments, which naturally come with embodiment mismatches, while also providing ground truth datasets for most environments. The third paradigm is relevant in the preference-based \gls{il} setting, where preference-ranked expert datasets are used to deal with settings in which sub-optimal demonstrations are available \cite{taranovic2023}. Together with the environments and tasks, these three paradigms can be combined to define the complexity of the \gls{il} setting, allowing the user to decide on the difficulty. This makes our benchmark versatile and suitable for different \gls{il} algorithms ranging from the simplest methods -- such as behavioral cloning -- to state-of-the-art methods. We collected the data via the Qualisys motion capture system, positioning a comprehensive set of markers to enable the capture of whole-body movements. Subsequently, we used the marker-based datasets to calculate human joint kinematics. We performed this computation using the OpenSim software platform, using a comprehensive full-body model, adapted from the previous work ~\cite{zhao2021}. In contrast, we generated the datasets for the Unitree A1 using Model Predictive Control~(MPC).

%
\begin{figure}
    \centering
    \includegraphics[width=.95\textwidth]{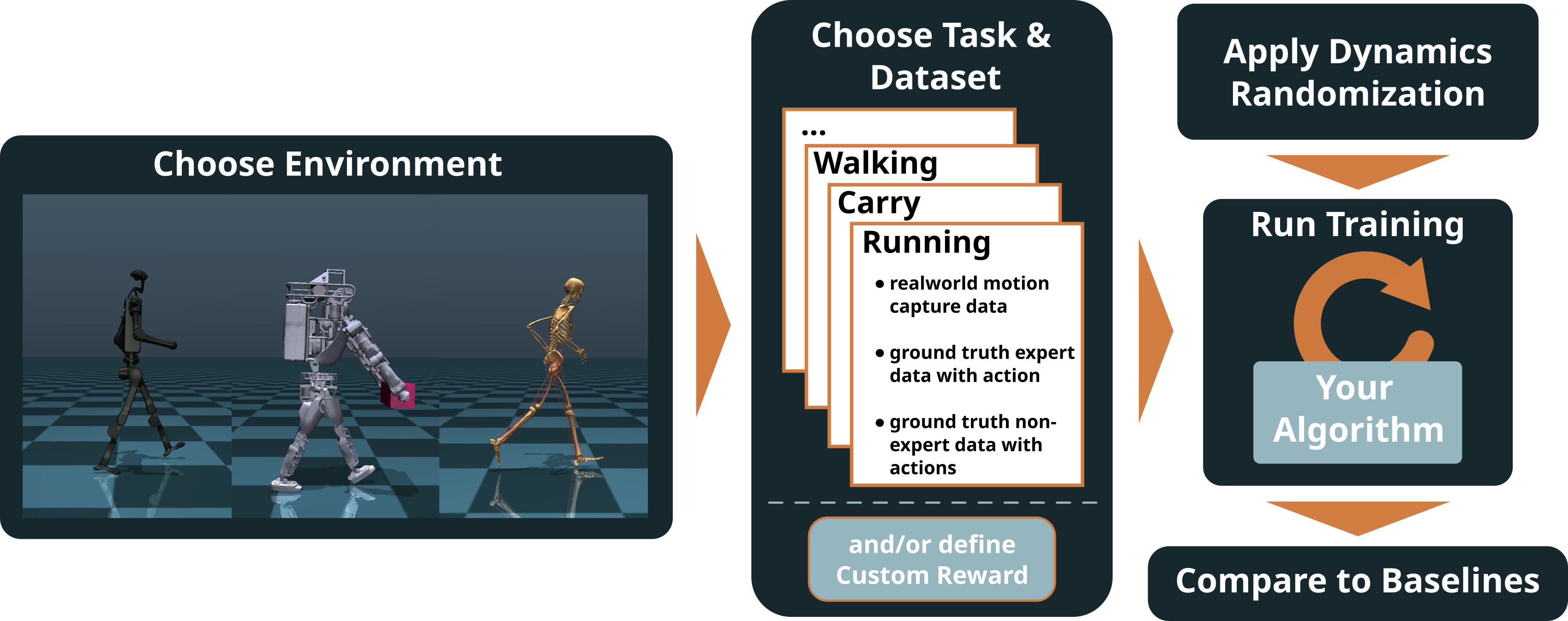}
    \caption{Training pipeline of LocoMuJoCo. First, an environment is chosen. Then, a task and dataset are chosen, and the training is started optionally with dynamics randomization. Finally, the performance of the algorithm can be compared to the expert performance or the performance of one of the provided baseline algorithms.}
    \label{fig:training_pipeline}
\end{figure}

\paragraph{Training.}
After choosing the task and dataset, the user has the option to define a dynamics randomization configuration. LocoMuJoCo supports randomization of different parameters, such as the properties of the joints, the inertia, or the friction between bodies. The user can choose whether to sample these parameters from a Gaussian with a provided standard deviation or from a uniform distribution with provided ranges. Furthermore, we provide simple handcrafted reward functions, which are based on the underlying goal of an expert's task. For example, we define the reward to be the difference in the center of mass velocity between the agent and the expert in the Talos walking task. Despite this reward being simple and not useful for \gls{rl} to train realistic gaits, it has proven very effective for evaluation in the \gls{il} setting. Alternatively, the use can define a custom reward function, making this benchmark also suitable for pure \gls{rl}. Finally, the user has the choice of implementing his algorithm in the framework of his choice by using the Gymnasium interface or by using Mushroom-RL. 

\paragraph{Evaluation.}
To show the feasibility of the benchmarks, we provide a set of baselines of classical \gls{irl} and adversarial~\gls{il} approaches.
We base all the implementations of these approaches on the Mushroom-RL library. For the Adversarial \gls{il} setting we provide \gls{gail}~\cite{Ho2016}, \gls{vail}~\cite{peng2019} and \gls{gaifo}~\cite{torabi2019_GaifO}. In the \gls{irl} setting we consider \gls{iq}~\cite{garg2021} and the \gls{lsiq}~\cite{alhafez2023} approaches. Furthermore, we provide the \gls{sqil}~\cite{reddy2020} algorithm, a simple ~\gls{il} approach that uses fixed rewards for expert and demonstration trajectories. The whole pipeline of our method is presented in Figure \ref{fig:training_pipeline}.

\section{Conclusion}
In this paper, we introduced LocoMuJoco, a novel benchmark for imitation learning in locomotion tasks. Addressing a notable gap in the field, our benchmark enables users to evaluate \gls{il} algorithms across various locomotion tasks, spanning a wide range of difficulty levels -- from relatively straightforward to highly challenging tasks.
Our benchmark covers most of the open problems in \gls{il}, including learning from real-world demonstrations or without expert actions. We provide fine-tuned baselines for most tasks using state-of-the-art \gls{il} algorithms. LocoMuJoco is easily extensible and provides straightforward interfaces to common \gls{rl} libraries, such as Gymnasium and Mushroom-RL. 
In future works, we aim to address a prevalent challenge in the majority of imitation learning (\gls{il}) benchmarks -- namely, devising effective methods for quantifying the quality of the cloned behavior. The current implementation provides a simple reward function measuring the overall gait quality. Our experiments show that a higher reward corresponds to a better imitation of the learned gait. Yet, deriving metrics grounded in biomechanical principles or in theoretical principles that account for the divergence between probability distributions is essential to propel research in this field.



\clearpage
\bibliography{library}
\bibliographystyle{plain}

\end{document}